\documentclass[10pt,twocolumn,letterpaper]{article}

\usepackage[]{cvpr}      % To produce the REVIEW version
\usepackage{times}
\usepackage[accsupp]{axessibility}  % Improves PDF readability for those with disabilities.
\usepackage[utf8]{inputenc} % allow utf-8 input
\usepackage[T1]{fontenc}    % use 8-bit T1 fonts
\usepackage{url}            % simple URL typesetting
\usepackage{booktabs}       % professional-quality tables
\usepackage{amsfonts}       % blackboard math symbols
\usepackage{nicefrac}       % compact symbols for 1/2, etc.
\usepackage{microtype}      % microtypography
\usepackage{xcolor}         % colors

\usepackage{graphicx}
\usepackage{amsmath,amssymb}
\usepackage{verbatim}
\usepackage{multirow}
\usepackage{marvosym}
\usepackage{makecell}
\usepackage{bm}
\usepackage{wrapfig}
\usepackage{caption}
\usepackage{twemojis}
\usepackage{utfsym}

\usepackage[pagebackref=true,breaklinks=true,letterpaper=true,colorlinks,bookmarks=false]{hyperref}

\usepackage{caption}
\usepackage{enumitem}
\setitemize{noitemsep,topsep=0pt,parsep=0pt,partopsep=0pt}

\definecolor{cvprblue}{rgb}{0.21,0.49,0.74}

\def\paperID{8291} % *** Enter the Paper ID here
\def\confName{CVPR}
\def\confYear{2025}

\title{
PartRM: Modeling Part-Level Dynamics with Large Cross-State Reconstruction Model
}

\let\oldtwocolumn\twocolumn
\renewcommand\twocolumn[1][]{
    \oldtwocolumn[{#1}{
    \centering
    \vspace{-10pt}
\includegraphics[width=0.85\textwidth]{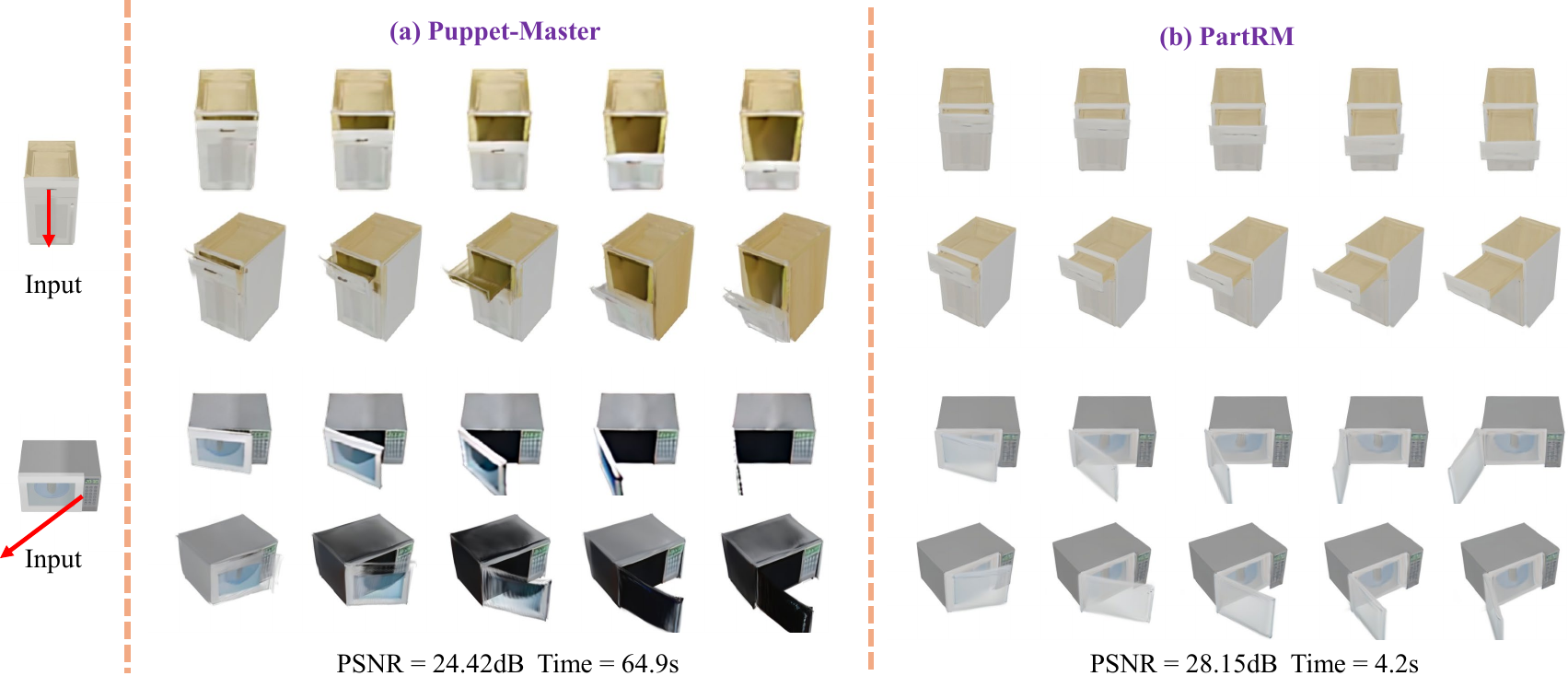}
\captionof{figure}{We present PartRM, given a \textbf{single-view} image and user-specific drags, PartRM can efficiently models appearance, geometry, and part-level motion in a feed-forward manner. Unlike the previous state-of-the-art, Puppet-Master \cite{li2024puppet}, PartRM achieves higher PSNR and significantly faster inference times. As shown in (b) compared to (a), PartRM produces 3D-aware, part-level motions with enhanced multi-view consistency, delivering more realistic and coherent results across different viewpoints.} %The inference time of \textcolor{red}{64.9s} refers to the process where we first apply drag deformation in 2D space and then leverage LGM \cite{tang2025lgm} for 3D reconstruction.}
\vspace{0.2cm}
        \label{fig:teaser}
    }]
}

\author{
Mingju Gao\textsuperscript{*1},  
Yike Pan\textsuperscript{*1,2},  
Huan-ang Gao\textsuperscript{*1,4}, 
Zongzheng Zhang\textsuperscript{1},  
Wenyi Li\textsuperscript{1},  \\
Hao Dong\textsuperscript{3},  
Hao Tang\textsuperscript{3}, 
Li Yi\textsuperscript{1},
Hao Zhao\textsuperscript{\textdagger 1,4}\\
\textsuperscript{1}Tsinghua University
\textsuperscript{2}University of Michigan 
\textsuperscript{3}Peking University
\textsuperscript{4}BAAI \\
Project Page: \url{https://PartRM.c7w.tech/}
}

\begin{document}

\maketitle
\setcounter{footnote}{1}
% % \renewcommand{\thefootnote}{}
\footnotetext{\textsuperscript{*}Equal Contribution. \textsuperscript{\textdagger}Corresponding Author.}
% % \footnotetext{Project page: \url{https://PartRM.c7w.tech}}
% % \tableofcontents

\begin{abstract}
As interest grows in world models that predict future states from current observations and actions, accurately modeling part-level dynamics has become increasingly relevant for various applications. Existing approaches, such as Puppet-Master, rely on fine-tuning large-scale pre-trained video diffusion models, which are impractical for real-world use due to the limitations of 2D video representation and slow processing times. To overcome these challenges, we present PartRM, a novel 4D reconstruction framework that simultaneously models appearance, geometry, and part-level motion from multi-view images of a static object. PartRM builds upon large 3D Gaussian reconstruction models, leveraging their extensive knowledge of appearance and geometry in static objects. To address data scarcity in 4D, we introduce the PartDrag-4D dataset, providing multi-view observations of part-level dynamics across over 20,000 states. We enhance the model’s understanding of interaction conditions with a multi-scale drag embedding module that captures dynamics at varying granularities. To prevent catastrophic forgetting during fine-tuning, we implement a two-stage training process that focuses sequentially on motion and appearance learning. Experimental results show that PartRM establishes a new state-of-the-art in part-level motion learning and can be applied in manipulation tasks in robotics. Our code, data, and models are publicly available to facilitate future research.
\end{abstract}

% \vspace{-6ex}
\vspace{-0.4cm}
\section{Introduction}
% \label{sec:intro}

% Modeling of dragging-based manipulation is important
World models are essential for predicting future states based on current observations and actions, enabling machines to understand and interact with the physical world. They play a crucial role in a variety of applications, including robotics \cite{wu2023daydreamer, zhou2024robodreamer, mendonca2023structured, ding2024preafford}, AR/VR, and beyond. Recently, there has been a growing interest in modeling part-level dynamics—the ability to generate realistic, fine-grained motion at the part level that accurately reflects current observations and user-specified drag interactions (see Figure \ref{fig:teaser}). This capability is key for tasks that demand high precision and adaptability, like manipulation and navigation in dynamic environments.

% generative modeling. puppet master. however, slow, view inconsistencies not reasoning about 3D, not ready for usage
However, our investigation into this promising field revealed that current part-level modeling approaches remain far from practical application. The state-of-the-art method, Puppet-Master \cite{li2024puppet}, fine-tunes a large pre-trained video diffusion model by adding a new conditioning branch to incorporate drag controls (see Figure \ref{fig:teaser} (a)). While this approach effectively leverages the rich motion patterns learned during pre-training, it falls short for real-world use. One major limitation is that it produces only single-view video as the output representation, whereas simulators require 3D representations to render scenes from multiple viewpoints. 
To meet the input requirements of the simulators, it is necessary for users to employ large reconstruction models based on monocular images. However, this approach may introduce additional sources of error.
Additionally, the diffusion denoising process can take several minutes to simulate a single drag interaction, which is counterproductive to the goal of providing rapid, trial-and-error feedback for generating manipulation policies.
% Puppet-Master fails to learn accurate part motions and does not ensure multi-view or temporal consistency. \eric{We conjecture?} We conjecture this limitation arises because it does not explicitly~\eric{explicitly} incorporate 3D information during the drag deformation process. Additionally, to synthesis novel views for Puppet-Master to do 3D reconstruction, the current novel view generation model (i.e., Zero123++ \cite{shi2023zero123++}) introduces artifacts in the generated views, which negatively impacts overall performance.~\eric{the logic is somehow broken from puppet-master to zero123}. Furthermore, the inference time for Puppet-Master generating a single image is approximately 64.9s~\eric{are you talking about puppet master?}, which is inefficient for large-scale data generation.

\noindent\textbf{Departure to 3D Representations.}
% To enable fast 3D reconstruction from input observation (i.e., images), we resort to large reconstruction models [ ] based on 3D Gaussian Splatting (3DGS) [ ], which has revonlutionzed the way of 3D reconstruction as they directly predict the content field from the input image in a feed-forward manner with scaled up parameter counts, reducing the recoustruction from several minutes to the level of seconds.
% However, different from them which 学习通用3d prior for reconstruction, we 建模图片中part的3d motion prior进而实现通用的simultanous 4D reconstruction of appearance and geometry, and motion learning (multiview consistent 3d partdragging).
% Although that seems hard, we point out that 同时刻画motion与geometry是有synergy的，because part-level motions are highly related to the geometry of the part (e.g., a 抽屉柜子一般沿着normal方向拉开）.
To enable rapid 3D reconstruction from input observations (i.e., images), we leverage large-scale reconstruction models based on 3D Gaussian Splatting (3DGS) \cite{kerbl3Dgaussians}.
% ~\eric{what is the connection between part-level dynamics modeling and 3D reconstruction? Why 3D reconstruction is needed to improve upon puppet-master? there should be a bridge here.} 
These models predict the content field from input images in a feed-forward manner, allowing them to reduce reconstruction time from several minutes (as in traditional optimization-based methods \cite{kerbl3Dgaussians, mildenhall2021nerf}) to just a few seconds.
While these models provide a robust 3D prior for generic reconstruction, our approach extends this capability by modeling the 3D motion priors of object parts. Specifically, we address the problem of simultaneous 4D reconstruction of appearance, geometry, and motion of parts from images.
% ~\eric{I feel the problem setup should be given more clearly earlier, involving input/output. I don't think everyone is familiar with the drag input but it is already used as everyone knows it.}
We posit that jointly modeling motion and geometry is essential, as part-level motions are inherently linked to the geometry of each part (e.g., a drawer typically slides along its normal direction when opened). This integration allows us to achieve a more realistic and interpretable representation of part-level dynamics.

\noindent\textbf{Challenges.}
Despite the clear motivation, developing a robust 4D reconstruction model presents significant challenges. A primary obstacle is the scarcity of accessible data capturing 3D objects along with their dynamic properties, which hinders the data-intensive training requirements of foundational models.
From a modeling standpoint, effectively representing drag interactions and incorporating these conditions into a 4D framework also remains an open challenge. Additionally, while we can leverage the pretrained capabilities of large reconstruction models \cite{tang2025lgm}—specifically their appearance and geometry modeling for static 3D objects—preserving these abilities during fine-tuning without catastrophic forgetting is another unresolved issue.
% ~\eric{the ``fine-tuning'' appears all of a sudden and I don't get the rationality why it is needed and what it is used for... }

\begin{figure}
\centering
\includegraphics[width=0.85\columnwidth]{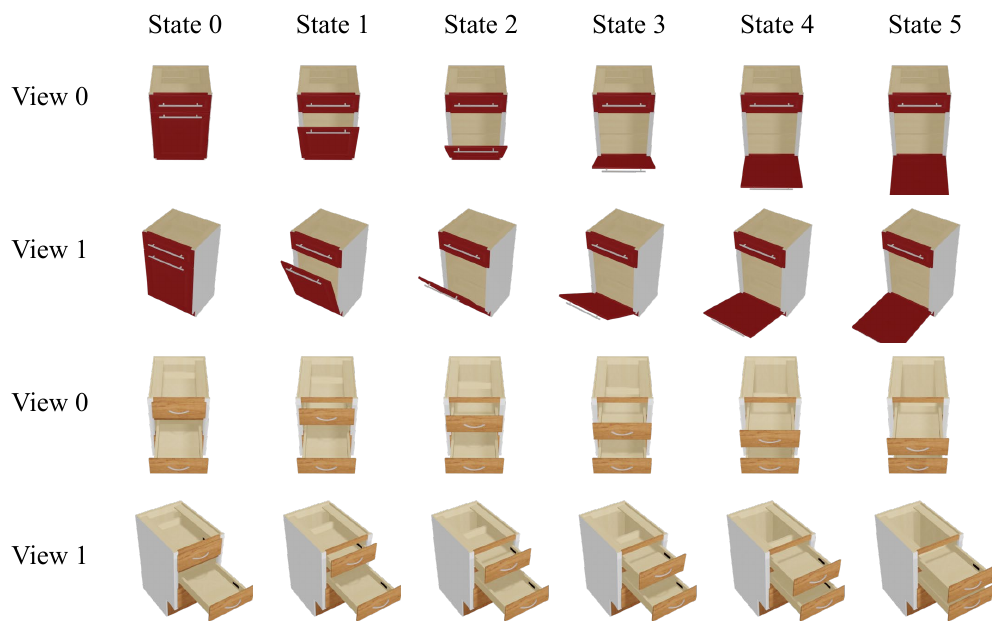}
\caption{\textbf{Introduction of PartDrag-4D dataset}. PartDrag-4D utilizes \textbf{738} meshes spanning 8 categories to generate \textbf{20,548} articulation states. For each state, PartDrag-4D renders 12 views. The drags are sampled on the moving surface.}
\vspace{-0.7cm}
\label{fig:dataset}
\end{figure}

In this paper, we present \textbf{PartRM}, a novel 4D reconstruction framework that simultaneously models the appearance, geometry, and part-level dynamics of objects.
To tackle the issue of data scarcity, we introduce the PartDrag-4D dataset (see Figure \ref{fig:dataset}), built on the PartNet-Mobility dataset \cite{xiang2020sapien}. We define a part's state by the extent of its movement within defined limits, rendering 738 objects with their parts in various positions. This setup provides multi-view observations across more than 20,000 distinct states.
To incorporate drag interactions, we propose a multi-scale drag embedding module that enhances the network’s capacity to recognize and process drag motions at multiple granularities.
To prevent catastrophic forgetting of pretrained appearance and geometry modeling capabilities, we introduce a two-stage training strategy. The first stage focuses on motion learning, supervised by matched 3DGS parameters, while the second stage focuses on appearance learning, using photometric loss to align rendered images with actual observations.
As illustrated in Figure \ref{fig:teaser}, our PartRM framework outperforms Puppet-Master, achieving higher PSNR and faster inference times. Additionally, it maintains both temporal and multi-view consistency under varying drag forces, demonstrating the effectiveness of our approach.

In our experiments, we established a new benchmark for part-level motion learning through novel view synthesis. We re-evaluated all existing methods \cite{li2025dragapart, li2024puppet, mou2024diffeditor} on this benchmark and achieved state-of-the-art results, setting a new standard in the field. 
% Additionally, we demonstrated that by reconstructing the same object in various states, our approach enables effective training of manipulation policies on synthetic data, showcasing its utility and robustness for real-world applications.
Furthermore, we demonstrated that by reconstructing the same object in multiple states, our approach can be applied to robotic manipulation tasks.
Our contribution can be summarized as follows: \begin{itemize}
    \item We introduce PartRM, a novel 4D reconstruction framework that simultaneously models appearance, geometry, and part-level motion.
    \item To address data scarcity, we develop the PartDrag-4D dataset, offering an extensive collection of multi-view observations of part-level dynamics. 
    \item We propose key innovations, including a multi-scale drag embedding module and a two-stage training approach, which enhance model performance \
    and mitigate catastrophic forgetting.
    \item Our approach achieves \textbf{state-of-the-art} results on newly established benchmarks for part-level motion learning and demonstrates its practical utility in real-world embodied scenarios.
\end{itemize}

\section{Related Work}

\textbf{Drag-conditioned Image \& Video Synthesis.}
Controlling image and video generation via dragging, which can be seen as a model of the dynamic world, enables simulation for robotics and autonomous systems. Methods for this task can be divided into training-free and training-based approaches.
Training-free methods iteratively update the source image to match user-specified drags, leveraging the power of diffusion modeling. 
% DragAPart, PuppetMASTER
For example, DragGAN \cite{pan2023drag} optimizes a latent representation of the image in StyleGAN \cite{karras2019style} to match user-specified drags. DragDiffusion adapts this idea to Stable Diffusion (SD) \cite{rombach2022high}. DragonDiffusion \cite{mou2023dragondiffusion} and Motion Guidance \cite{gengmotion} combine SD with a guidance term that captures feature correspondences and a flow loss, and DiffEditor \cite{mou2024diffeditor} combines stochastic differential equation (SDE) into ordinary differential equation (ODE) sampling to increase flexibility while maintaining content consistency over DragonDiffusion.
Training-based methods, on the other hand, learn drag-based control using specifically designed training data, mostly leveraging powerful generative modeling capabilities of diffusion models \cite{blattmann2023stable,rombach2022high,zhang2023adding,gao2024scp,li2024fairdiff,zhang2024ctrl,ni2025straight,chen2024ultraman,xu2024diffusion,li2025avd2}. For example, iPoke \cite{blattmann2021ipoke} trains a variational autoencoder (VAE) to synthesize videos with objects in motion. MCDiff \cite{chen2023motion} and YODA \cite{moser2023yoda} train diffusion models using DDPM and flow matching, respectively. Li et al. \cite{li2024generative} employ a Fourier-based representation of motion suitable for natural, oscillatory dynamics characteristic of objects like trees and candles, generating motion with a diffusion model. DragNUWA \cite{yin2023dragnuwa} and MotionCtrl \cite{wang2024motionctrl} extend text-to-video generators with drag-based control.
DragAPart \cite{li2025dragapart} and Puppet-Master \cite{li2024puppet} propose learning a generalist motion model but focus solely on single-view images and videos, respectively. In contrast, our work utilizes 3D Gaussians as the state representation, advancing towards the simulation of robotics policies.

\noindent\textbf{Large Reconstruction Models.}
% citation from l4gm
Large Reconstruction Models (LRMs) replace the costly optimization-based 3D representation learning with feedforward approaches by training neural networks to directly regress the 3D representation, especially implicit representation~\cite{zhou2023pad, zhong2022snake, yuan2024slimmerf, chen2023nerrf,liu2024rip,chen2024rgm}. LRM \cite{honglrm} was among the first to utilize large-scale multiview datasets, including Objaverse \cite{deitke2023objaverse}, to train a transformer-based model for NeRF reconstruction. The resulting model demonstrates better generalization and higher-quality reconstruction of object-centric 3D shapes from sparse posed images in a single model forward pass. Similar works have explored changing the representation to Gaussian splatting \cite{tang2025lgm, zhang2025gs, jiang2024brightdreamer, Zou_2024_CVPR, song2024sa}, introducing architectural changes to support higher resolution \cite{xu2024grm, shen2024gamba}, and extending the approach to 3D scenes \cite{charatan2024pixelsplat, chen2025mvsplat, zheng2024gps, tian2024drivingforward}. These methods can be generalized to input images supplied from sampling multiview diffusion models, allowing fast 3D generation \cite{li2023instant}.
L4GM \cite{ren2024l4gm} produces dynamic 3D representations from single-view video input by training on rendered animated objects in Objaverse. However, it is not action-conditioned, which means it does not function as a world model. Additionally, it is only trained on simple animations and does not support part-level dynamics.
Instead, in this paper, we introduce PartRM which achieves simultaneous appearance, geometry, and part-level motion modeling.

\section{The Proposed PartDrag-4D Dataset}
\label{sec:our_data}

\begin{figure*}
\centering
\includegraphics[width=0.85\textwidth]{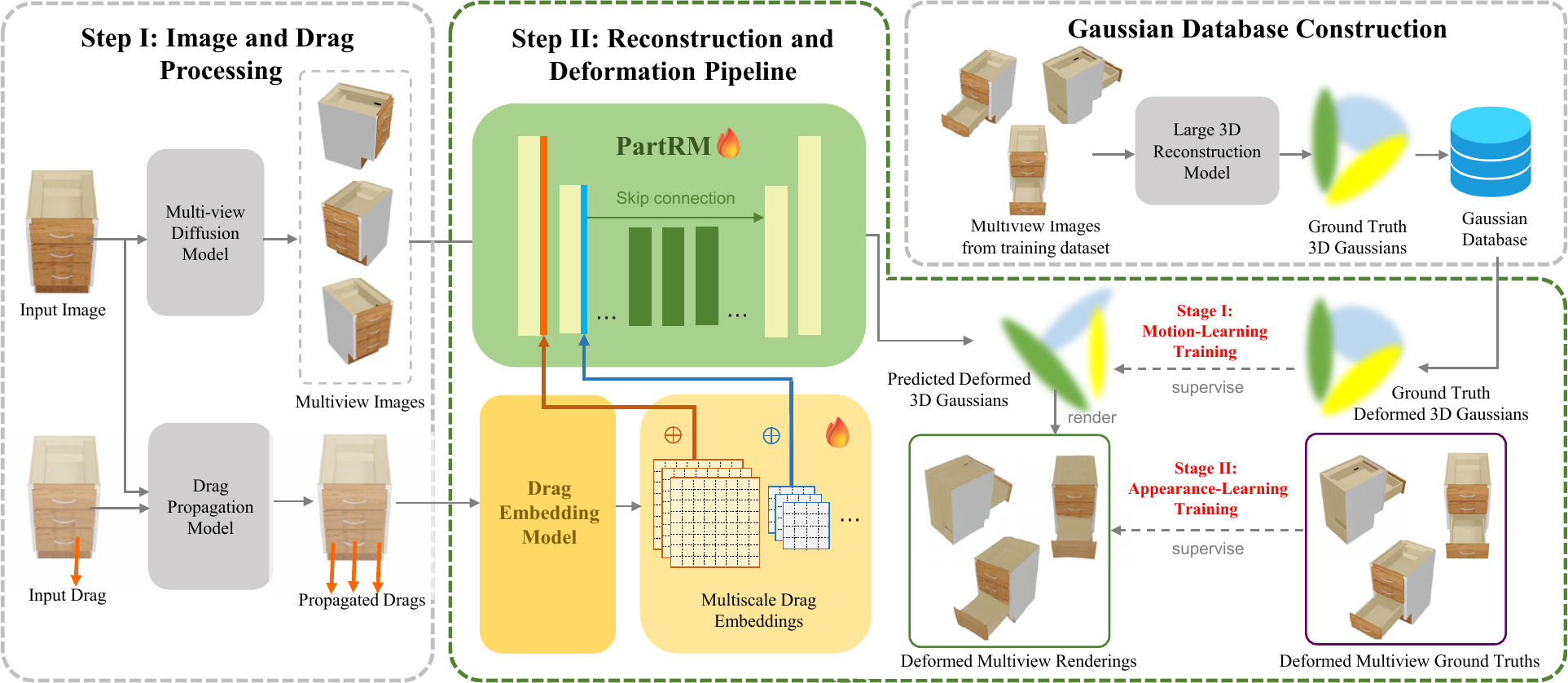}
\caption{\textbf{Overview of PartRM.} We first leverage a fine-tuned Zero123++ to generate multi-view images, followed by our designed drag propagation module to distribute drags on the moving parts. The drags and multi-view images are then fed into our designed network, where the drags are embedded using our multi-scale embedding module and subsequently concatenate  to the UNet down blocks. We adopt a two-stage training approach: in the first stage, the network learns part motion using ground truth deformed 3D Gaussians as supervision, which are stored in the Gaussian database constructed by LGM. In the second stage, the network learns appearance, with ground truth deformed multi-view renderings serving as supervision.
}
\vspace{-0.5cm}
\label{fig:main}
\end{figure*}

\textbf{Formulation. }We begin by defining our notations. Let \( o_t \) and \( s_t \) represent the observation and state of the object at time \( t \), respectively. In the context of part-level dynamic learning, \( o_t \) denotes the 2D rendering of the current state \( s_t \), and \( s_t \) represents the pose of each part, containing the 3D information necessary to render \( o_t \). 
Let \( a_t \) denote the action applied to the object. 
Our objective is to develop a simultaneous model for appearance, geometry, and part-level dynamics using a large 4D reconstruction model \( f \), which can be expressed as,
\begin{equation}
\setlength\abovedisplayskip{3pt}
\setlength\belowdisplayskip{3pt}
    \label{eq:world_model}
    f: (o_t, a_t) \rightarrow s_{t+1}.
\end{equation}
This model aims to predict the state \( s_{t+1} \) of the object in the next time step given the current observation \( o_t \) and the action~\( a_t \).

\noindent\textbf{Motivation.}
To train such a model, it is imperative to collect data pairs that fully meet the requirements of our task, as detailed above.
Our investigation into this field reveals two primary streams of datasets.
One stream consists of \((o_t, a_t, o_{t+1})\), which includes only paired images with their corresponding drag information \cite{li2025dragapart}, but lacks essential 3D data that can be utilized for further simulation and application.
The other stream utilizes a general dataset, such as Objaverse, filtering out static and drastically changing objects, and annotates drag by sampling points from the entire surface of the object and describing the movement of these points.
However, we argue that this method of generation does not conform to kinematic dynamics.
Data constructed in this manner not only involves movement but also includes deformation and other operations, which simultaneously alter both shape and appearance.
% \vspace{-1ex}

\noindent\textbf{PartDrag-4D Dataset.}
Our dataset is constructed using PartNet-Mobility \cite{xiang2020sapien}, which provides detailed part-level annotations for articulated objects. We utilize 738 meshes spanning 8 categories (using 7 categories for training and all 8 for evaluation). For each mesh, as illustrated in Figure \ref{fig:dataset}, we animate one articulated part through 6 stages between two extreme positions (e.g., a drawer fully opened and fully closed), while setting the other parts to random positions, yielding a total of 20,548 states (a mesh may have multiple movable parts), with 20,057 states for training and 491 states for evaluation. Throughout the animation sequence, all the other parts remain in fixed positions. We store all the animated meshes, point clouds, and corresponding moving points in our dataset for subsequent stages.

To render multi-view images of each state of the 3D mesh, we use Blender to generate a total of 12 views with a fixed camera distance. For drag sampling on the surface of the moving parts, we project the sampled 3D points into the 2D image space using the given camera parameters. 
To ensure that the projected points correspond accurately to the visible surface of the mesh from the specified camera perspective, rather than residing within the interior of a part, we remove points from the projected 2D drags that exhibit depths significantly greater than those of their neighboring points, which may be occluded by other points.
% \vspace*{-15pt}
% A critical component of our training process is the development of an appropriate 3D dataset that includes drag annotations. Such a dataset must consist of triplets $(x, y, \mathcal{D})$, where $\mathcal{D}$ represents the set of 3D drag operations applied to the 3D mesh $x$, resulting in a deformed 3D mesh $y$. However, existing datasets only contains only paired images with their corresponding drag information, but lacks essential 3D data. Therefore, to accomplish our objectives, it is necessary to construct a specialized dataset that fully accommodates the requirements of our task.

% \vspace{-1.5ex}
\vspace*{-5pt}
\section{The Proposed Method}
\subsection{Overview}

We provide an overview of our pipeline in Figure~\ref{fig:main}. Given a single-view observation \(o_t \in \mathbb{R}^{h \times w \times 3}\) and input drags \(a_t\), where \(a_t\) is parameterized by its start and end points projected on \(o_t\), i.e., \(a_t = (a_{t,\text{src}}(x,y), a_{t,\text{dst}}(x,y))\), our objective is to generate a 3D Gaussian representation \(s_{t+1}\) that depicts the state $s_{t+1}$ after the dragging. The challenge lies in simultaneously conducting appearance and geometry modeling (\(o_t \rightarrow s_t\)) for 3D and motion learning (\(s_t, a_t \rightarrow s_{t+1}\)) in part level.
To ensure the reconstruction model accurately identifies which parts to move and how to move them, we have developed a drag-propagation module to distribute the input drags across the moving parts, as detailed in Sec.~\ref{sec:pre-process}. After propagating the drag condition, we delve into the embedding of these conditions in Sec.~\ref{sec:drag_embedding}.
For efficient training while avoiding catastrophic forgetting of pretrained knowledge of appearance and geometry modeling for static 3D objects, we implement a two-stage training pipeline, detailed in Sec.~\ref{sec:2-stage}. The first stage focuses on learning the motion dynamics, while the second stage is dedicated to learning the appearance characteristics.

% Given a single-view image $I \in \mathcal{R}^{3 \times H \times W}$ and input drags $\mathcal{D}=\{d_k\}_{k=1}^K$, where $d_k = (s_k, t_k)$ and $s_k = (s_{kx}, s_{ky}), t_k = (t_{kx}, t_{ky})$, our objective is to generate deformed 3D Gaussian representations of the modified image according to the specified input drags. As illustrated in Figure \ref{fig:main}, we first employ a diffusion model to synthesize multi-view images from $I$. To more effectively leverage the characteristics of articulated objects, we have developed a drag-propagation module to distribute the input drags across the moving parts during the inference of our articulated dataset. The generated images and drags are then processed by our meticulously designed model, which encodes the drag conditions into the reconstruction process to yield the expected deformed 3D Gaussian representations.

% We implement a two-stage training pipeline, where the first stage is focused on learning the motion dynamics, while the second stage is dedicated to learning the appearance characteristics. At the first training stage, we creatively leverage ground truth Gaussians from our pre-built database as the supervision signals to facilitate fast and effective convergence.

% \subsection{Image and Drag Preprocessing}
\subsection{Image and Drag Preprocessing}
\label{sec:pre-process}
\textbf{Multi-view Images Generation.} We build PartRM upon LGM \cite{tang2025lgm}, which requires multi-view images as input. To generate these images, we utilize Zero123++ \cite{shi2023zero123++} due to its strong multi-view consistency. To further enhance the quality of the generated multi-view images, we fine-tune Zero123++ using our training dataset \cite{xu2024instantmesh}. The novel-view images, in conjunction with the original image, are fed into the PartRM network. This fine-tuning process optimizes the model's performance, ensuring the production of high-quality multi-view representations for subsequent processing.

\begin{figure}
\centering
\includegraphics[width=0.8\columnwidth]{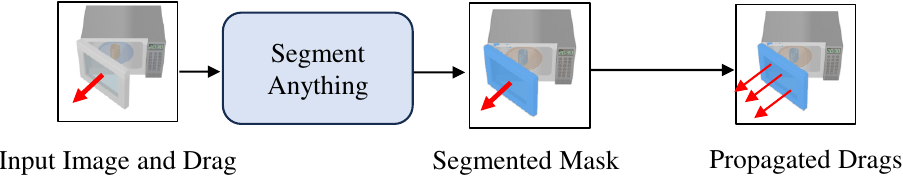}
\caption{\textbf{Illustration of drag propagation module.}}
\vspace{-0.6cm}
\label{fig:propagation}
\end{figure}

\noindent\textbf{Drag Propagation.}
Providing a large reconstruction model with a single dragging condition can lead to hallucinations due to the inherent ambiguity in a single drag. At this stage, we leverage kinematic priors of articulation motion to augment the input drag condition.
% To optimize the utilization of the drag motion associated with the movement of articulated objects, we propose a drag propagation module designed to accurately distribute drag across the surface of the moving part. 
As illustrated in Figure \ref{fig:propagation}, for each input pair $(o_t, a_t)$, our drag propagation module first processes the input image and utilizes the drag start point to query the Segment Anything model \cite{kirillov2023segment}. This model generates a part segmentation mask, which delineates the area for drag propagation. This approach generates the distribution of drag proposals that are appropriately confined to the relevant regions of the articulated part, enhancing the accuracy and efficiency of the motion representation. 

After this, we sample points on the part segmentation mask as the starting points of the propagated drags and use the same drag intensity (i.e., the magnitude of relative position change) as the input drag. We denote $\Delta a_t {=} a_{t,\text{dst}} {-} a_{t,\text{src}}$. The \(i\)-th propagated drag \(a_{t,i}\) of the input drag \(a_t\) can be formulated as:
\begin{equation}
\setlength\abovedisplayskip{3pt}
\setlength\belowdisplayskip{3pt}
    a_{t,i} = (a_{t,i,\text{src}}, a_{t,i,\text{src}} + \Delta a_t),
\end{equation}
where \(a_{t,i,\text{src}}\) is the \(i\)-th point sampled from the part segmentation mask. 
Despite the potential inaccuracy in the estimation of the magnitude of drag intensity, our subsequent model remains robust enough to learn to generate the expected output in a data-driven manner.

\subsection{Drag Embedding}
\label{sec:drag_embedding}

% LGM
The U-Net of PartRM, which follows LGM \cite{tang2025lgm}, is constructed with residual layers \cite{he2016deep} and self-attention layers \cite{vaswani2017attention}, similar to previous works \cite{ho2020denoising, metzer2023latent, szymanowicz2024splatter}. It is designed to be asymmetric, with a smaller output resolution compared to the input, allowing for higher-resolution input images while limiting the number of output Gaussians. In this section, we embed the propagated drags of the input views \(\{a_{t,i}\}_{i=1}^N\) into multi-scale drag maps and interact them with each downsample block of the U-Net. This enhances the network's capacity to recognize and account for various granularities of drag motion.

\begin{figure}
\centering
\includegraphics[width=0.85\columnwidth]{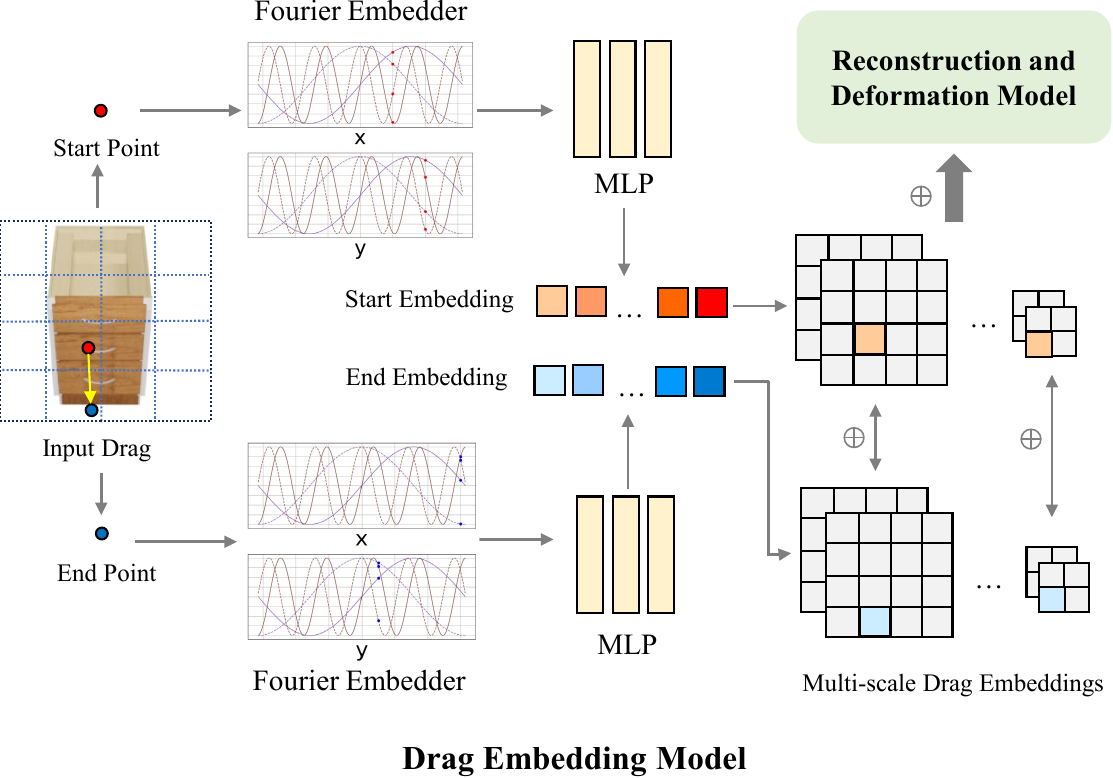}
\caption{\textbf{Illustration of drag embedding module.}}
\vspace{-0.6cm}
\label{fig:embedding}
\end{figure}

% Take multiview images and propagated drags as inputs, we utilize a 3D-UNet\cite{todo UNet} to do the reconstruction and deformation. To make the network aware of the input drags, we propose a multi-scale drag embedding module to enhance the network's capacity to recognize and account for various levels of drag motion.
As illustrated in Figures \ref{fig:main} and \ref{fig:embedding}, we pre-compute multi-scale drag maps with respect to the spatial dimensions of the U-Net's down-sample block outputs. For the output \(O_l \in \mathbb{R}^{C_l \times H_l \times W_l}\) of the \(l\)-th block \(D_l\), and for each drag \(\{a_{t,i}\}_{i=0}^N\), where \(i=0\) denotes the input drag and \(1 \le i \le N\) denotes the propagated drags, we first encode the coordinates of each starting point \(a_{t,i,\text{src}}\) and ending point \(a_{t,i,\text{dst}}\) with a Fourier embedder and a 3-layer MLP to obtain embeddings for the start point \(F(a_{t,i,\text{src}})\) and the end point \(F(a_{t,i,\text{dst}})\). Then the drag map \(M_{t,l,i} \in \mathbb{R}^{C_M \times H_l \times W_l}\) is defined as:
\begin{equation}
\setlength\abovedisplayskip{3pt}
\setlength\belowdisplayskip{3pt}M_{t,l,i}\left[a_{t,i,\text{src}}\right] = F(a_{t,i,\text{src}}) \oplus F(a_{t,i,\text{dst}}),
\end{equation}
% \begin{equation}
%     F(a_{t,i,\text{src},x}, a_{t,i,\text{src},y}) = \text{FC}(\text{Fourier}(a_{t,i,\text{src},x}, a_{t,i,\text{src},y}))
% \end{equation}
% \begin{equation}
%     F(a_{t,i,\text{dst},x}, a_{t,i,\text{dst},y}) = \text{FC}(\text{Fourier}(a_{t,i,\text{dst},x}, a_{t,i,\text{dst},y}))
% \end{equation}
% \begin{equation}
%     M_{t,l,i}\left[a_t,i,src\right] = F(a_t,i,src)  \oplus  F(a_t,i,dst)
% \end{equation}
% As illustrated in Figure \ref{fig:main} and \ref{fig:embedding}, we pre-compute multi-scale drag maps w.r.t. the UNet down-sample block output's spatial dimensions. For the $l$-th block $D_l$ output $O_l \in \mathcal{R} ^ {C_l \times H_l \times W_l}$, for every drag $d_i = (s, t) \in \{d_k\}_{k=1}^{K} \cup \{p_k\}_{k=1}^{NK}$, where $N$ is the number of the propagated drags for each input drag, the drag map $M_{li} \in \mathcal{R}^{C_M \times H_l \times W_l}$ is defined as:
% \begin{equation}
%     F(s_{ix}, s_{iy}) = FC(Fourier(s_{ix}, s_{iy}))
% \end{equation}
% \begin{equation}
%     F(t_{ix}, t_{iy}) = FC(Fourier(t_{ix}, t_{iy}))
% \end{equation}
% \begin{equation}
%     M_{li}[\lfloor \frac{s_{ix}H_l}{H} \rfloor , \lfloor \frac{s_{iy}W_l}{W} \rfloor] =  F(s_{ix}, s_{iy}) \oplus  F(t_{ix}, t_{iy})
% \end{equation}
where $\oplus$ means concentration along the channel dimension. The other elements of $M_{t,l}$ are set to zero. To obtain $M_{t,l}$ for all drags, we sum all the $M_{t,l,i}$, which means $M_{t,l} = \sum_{i=0}^{N}M_{t,l,i}$.
After we get the drag map \(M_{t,l}\), we concatenate \(M_{t,l}\) and \(O_l\) along the channel dimension and feed them into a convolutional layer to match the channel dimensions for the input of the next down-sample block. The input of the \((l+1)\)-th down-sample block \(I_{l+1}\) is defined as:
\begin{equation}
\setlength\abovedisplayskip{3pt}
\setlength\belowdisplayskip{3pt}
    I_{l+1} = O_l + \text{Conv}(M_{t,l} \oplus O_l),
\end{equation}
where \(M_{t,l} \oplus O_l \in \mathbb{R}^{(C_M + C_l) \times H_l \times W_l}\), \(I_{l+1} \in \mathbb{R}^{C_l \times H_l \times W_l}\), and parameters of the convolution layer are initialized to zero.

\subsection{Two-stage Training Pipeline}
\label{sec:2-stage}

To retain the ability to model the appearance and geometry of static 3D objects, we build our PartRM upon pretrained LGM \cite{tang2025lgm}, which uses an asymmetric U-Net as a high-throughput backbone operating on multi-view images to generate high-resolution 3D Gaussians from multi-view images.
However, directly fine-tuning on the proposed dataset can result in catastrophic forgetting of previously learned knowledge, which impairs generalization when dealing with in-the-wild data.
To address this issue, we propose a two-stage learning method. First, we focus on learning motions, which have not been previously learned. Then, we conduct simultaneous training of appearance, geometry, and motion. This approach allows us to train in a coarse-to-fine manner, ensuring better retention and generalization capabilities.

\noindent\textbf{Motion Learning Stage.} In the initial training stage, our goal is to learn the motion induced by drag effects. We tackle this problem using a knowledge distillation (KD) approach. Specifically, we use Gaussians inferred from the target observations by the pre-trained network to supervise the training of the student network, which is provided with source state observations (see Gaussian Database Construction in Figure~\ref{fig:main}). Leveraging the output from the pre-trained network itself facilitates continual learning, which keeps the ability of generalization while speeds up the training process.

\begin{table*}[htbp]
    \centering
    \resizebox{0.9\linewidth}{!}{%
        \begin{tabular}{ccccccccc}
            \toprule
            \multirow{2}{*}{\textbf{Method}} & \multirow{2}{*}{\textbf{Setting}} & \multicolumn{3}{c}{\textbf{PartDrag-4D}} & \multicolumn{3}{c}{\textbf{Objaverse-Animation-HQ}} & \multirow{2}{*}{\textbf{Time} (↓)} \\ 
            \cmidrule(lr){3-5} \cmidrule(lr){6-8}
             & & \textbf{PSNR (↑)} & \textbf{SSIM (↑)} & \textbf{LPIPS (↓)} & \textbf{PSNR (↑)} & \textbf{SSIM (↑)} & \textbf{LPIPS (↓)} \\ 
            \midrule
            DiffEditor \cite{mou2024diffeditor} & NVS-First & 22.52 & 0.8974 & 0.1138  & 19.24 & 0.8988 & 0.0902 & 33.6s / 151.2s \\ 
            DiffEditor \cite{mou2024diffeditor} & Drag-First & 22.34 & 0.9174 & 0.0918  & \underline{19.46} & \underline{0.9079} & \underline{0.0842} & 11.5s / 128.8s \\ 
            DragAPart \cite{li2025dragapart} & NVS-First & 24.27 & 0.9343 & 0.0690  & 19.38 & 0.8915 & 0.0873 & 21.4s / 139.7s \\ 
            DragAPart 
            \cite{li2025dragapart} & Drag-First & \underline{24.91} & 0.9454 & 0.0567  & 19.44 & 0.9004 & 0.0885 & \underline{8.5s / 119.4s} \\ 
            Puppet-Master \cite{li2024puppet} & NVS-First & 24.20 & 0.9447 & 0.0579  & - & - & - & 64.9s / 187.5s \\ 
            Puppet-Master \cite{li2024puppet} & Drag-First & 24.42 & \underline{0.9475} & \underline{0.0528}  & - & - & - & 245.8s / 361.5s \\ 
            \textbf{PartRM (Ours)} & - & \textbf{28.15} & \textbf{0.9531} & \textbf{0.0356} &  \textbf{21.38} & \textbf{0.9209} & \textbf{0.0758} & \textbf{4.2s / -} \\ 
            \bottomrule
        \end{tabular}
    }
    \caption{\textbf{Comparison of our method and baseline methods on PartDrag-4D and Objaverse-Animation-HQ}. \textbf{NVS-First} refers to the approach where multi-view images are first generated, followed by drag deformation applied to each view. \textbf{Drag-First} means drag deformation is applied to the image first, and then multi-view images are generated based on the deformed image. The time column contains two values separated by a slash: the first value indicates the time spent on 3D reconstruction using LGM, and the second value represents the time spent on 3D reconstruction using optimization-based 3D Gaussian splatting. PartRM has only one value in the time column because it simultaneously models appearance, geometry, and part-level motion. PartRM achieve \textbf{State-of-the-Art} on all metrics.}
% \vspace{-0.6cm}
    \label{tab:sota}
    \vspace{-0.5cm}
\end{table*}

For matching between these two sets of Gaussians, intuitively we should perform matching according to the coordinates of these Gaussians. However, leveraging the advantages of using splatter images \cite{szymanowicz2024splatter} as representations, we can enforce the network to learn to transform the representation to the target state before the output regression layer.
In other words, we can directly apply L2 loss on the 14-dimensional parameters of the corresponding pixels in the 2D images. Formally, this can be written as:
\begin{equation}
\setlength\abovedisplayskip{3pt}
\setlength\belowdisplayskip{3pt}
    \mathcal{L}_1 = \sum_{i \in h \times w \times v, j=i} \left\| \mathcal{GS}_i - \mathcal{GS}_j \right\|_2^{2},
\end{equation}
where \(i\) denotes corresponding pixels in the splatter images, and \(\mathcal{GS}_i\) and \(\mathcal{GS}_j\) represent the 14-dimensional parameters of the Gaussians at those pixels.

% for each training instance，which comprises multi-view images and their corresponding camera parameters, we employ the fine-tuned LGM to predict the ground truth Gaussians, represented as splatter images. These predictions are then aggregated to form our Gaussian database, which will serve as a foundation for subsequent training.
% \textbf{Motion learning stage}: The initial training stage is designed to learn the motion induced by drags effects. For supervision, we extract the ground truth Gaussian representations from the pre-constructed database. Given that the output of the reconstruction 3D U-Net are splatter images \cite{todo splatter image}, we can view these outputs as feature maps in the dimension of $\mathcal{R}^{14*H*W}$. The output gaussians and the ground truth have a natural correspondence at the same spatial location in the splatter image. Therefore, we can formulate our loss function as:
% \begin{equation}
%     loss_{stage1} = \sum_{i=1}^{H}\sum_{j=1}^{W} (gt[i, j] - pred[i, j])^2
% \end{equation}
% In the absence of strong supervision, such as ground truth splatter images, if we only use multi-view images as supervision, the reconstruction model is unable to accurately capture the dynamics of drag. To investigate this, we perform an ablation study as part of our experimental analysis.
\noindent\textbf{Appearance Learning Stage.} After the motion learning stage, we introduce an additional stage to jointly optimize appearance, geometry, and part-level motion. Specifically, we replace the supervision signal from Gaussians with target observations. In each training step, we render the RGB image and the alpha image of eight novel views. For every view, we formulate our loss as:
\begin{equation}
\setlength\abovedisplayskip{3pt}
\setlength\belowdisplayskip{3pt}
\begin{aligned}
    \mathcal{L}_2 = L_{\text{mse}}(I_{\text{pred\_color}}, I_{\text{gt\_color}}) \\ 
    + \lambda_1 L_{\text{lpips}}(I_{\text{pred\_color}}, I_{\text{gt\_color}}) \\ 
    + \lambda_2 L_{\text{mse}}(I_{\text{pred\_alpha}}, I_{\text{gt\_alpha}}),
\end{aligned}
\end{equation}
where \(\lambda_1\) and \(\lambda_2\) are set to 1.0 during our training.

% If we only use splatter image as the supervision described as above, the model cannot model the appearance well. Therefore we devise a coarse-to-fine training stage to learn the appearance. At each training step, we render the RGB image and alpha image of eight views. For every view, we formulate our loss as:
We provide ablation studies on this coarse-to-fine training approach in Table \ref{tab:ablation_stage}. The results demonstrate that if we only use multi-view images as supervision, the reconstruction model fails to learn the part-level motion effectively. The 3D representation will merely attempt to exploit the loss function in stage 2 without learning the appropriate motion. 
% This proves the necessity of the motion learning stage.
% \vspace{-0.2cm}

% In the absence of strong supervision, such as ground truth splatter images, if we only use multi-view images as supervision, the reconstruction model is unable to accurately capture the dynamics of drag. To investigate this, we perform an ablation study as part of our experimental analysis.
% Due to the differing convergence paths of the two losses, we cannot complete the training in a single phase. We conduct the ablation study in our experiment.

% \vspace{-0.4cm}
\section{Experimental Results}

\begin{figure*}
\centering
\includegraphics[width=0.75\textwidth]{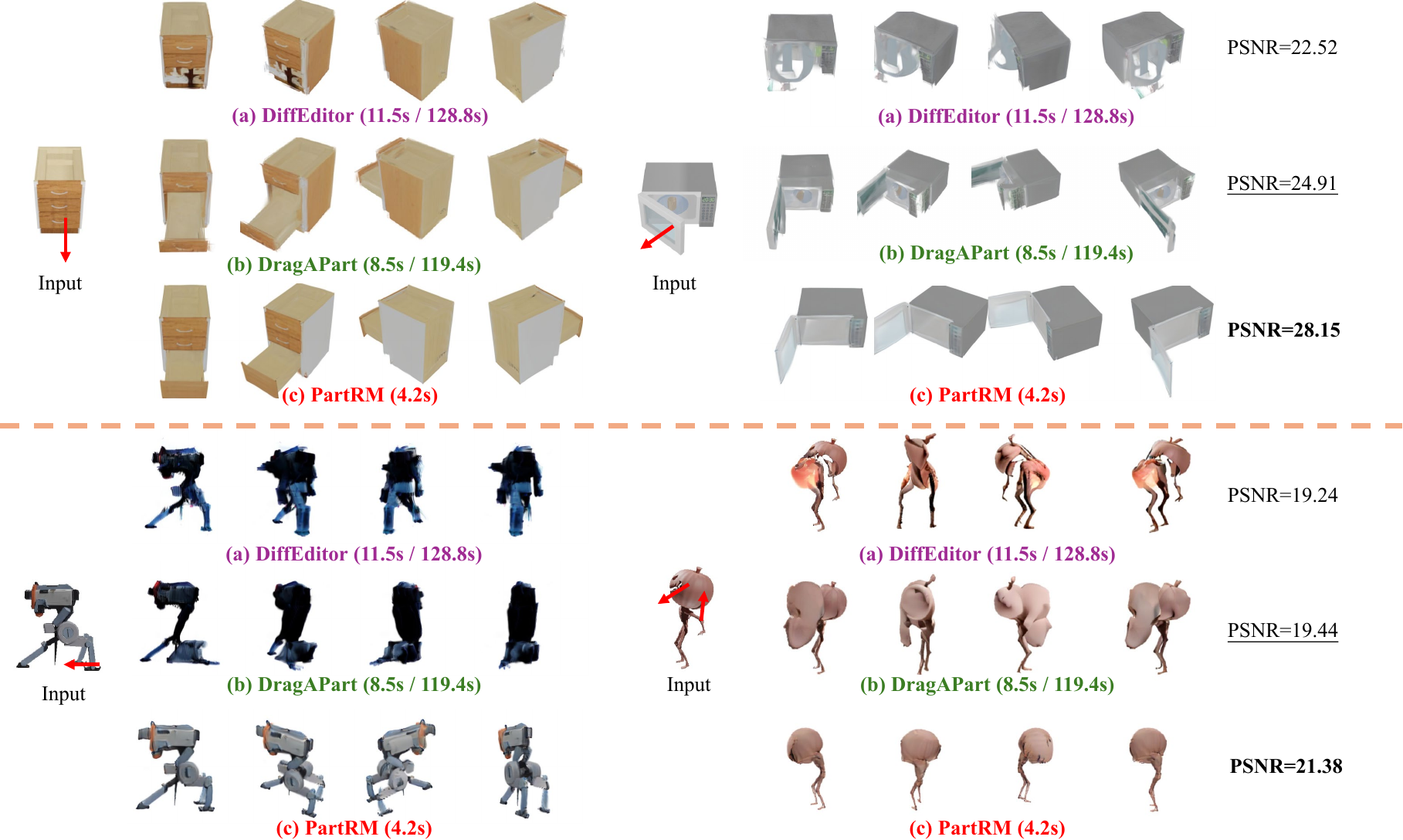}
\caption{\textbf{Qualitative comparisons between PartRM and baselines}. 
The time values separated by the slash represent the time spent first applying 2D drag deformation to the input image and then performing 3D reconstruction using LGM and the optimization-based method, respectively. PartRM has only one value in the time column because it simultaneously models appearance, geometry, and part-level motion, eliminating the need for separate steps. PartRM learn the part motion effectively.
}
\label{fig:quati}
\vspace{-0.4cm}
\end{figure*}

\subsection{Experiment Settings}

In addition to the PartDrag-4D dataset introduced in Sec.~\ref{sec:our_data}, we also evaluate our method on the Objaverse-Animation-HQ dataset \cite{li2024puppet}, which is derived from the animated data in the Objaverse \cite{deitke2023objaverse} dataset. Due to the large volume of data in the Objaverse-Animation-HQ dataset, we sample approximately 15,000 states and manually split them into training and evaluation sets. For evaluation, we render eight views of each predicted gaussian splatting and compute PSNR, SSIM, and LPIPS metrics between the ground truth images and the rendered images at a resolution of \(256 \times 256\).

% We evaluate our method on two challenging datasets: PartDrag-4D dataset which is mentioned in section \ref{sec:our_data} and Objaverse-Animation-HQ \cite{todo pm}, which is built on animated data in Objaverse \cite{todo ob} dataset. Since the tremendous data in Objaverse-Animation-HQ dataset, we sample about 15,000 states from it and split the training and evaluation data manually. For evaluation, we randomly render eight views of each generated Gaussian Splattings, and compute PSNR, SSIM and LPIPS between the ground images and rendered images at a resolution of $256\times 256$. All the training and inference are on the NVIDIA A800 80G GPUs.

\subsection{Main Results}
% Since the absence of prior research addressing 2D drag deformation to 3D drag-deformed Gaussian representations, we propose two potential approaches for aligning our work with existing methodologies. (i) First, applying drag deformation in 2D space, followed by 3D reconstruction to Gaussian Splattings. (ii) Alternatively, reconstructing the input image into 3D Gaussian splattings and performing drag deformation directly in 3D space. However, due to the lack of open-source implementations for the drag-based 3D gaussians deformation, we focus our comparisons with baseline methods solely on the first approach. 

\textbf{Baselines.} We conduct a comprehensive comparison of our method with state-of-the-art drag-based generative models, specifically \textcolor{ForestGreen}{DragApart} \cite{li2025dragapart}, \textcolor{purple}{DiffEditor} \cite{mou2024diffeditor}, and \textcolor{orange}{Puppet-Master} \cite{li2024puppet}. For the training-free model \textcolor{purple}{DiffEditor}, we utilize its official implementation and pre-trained checkpoint. To ensure a fair comparison with \textcolor{ForestGreen}{DragApart}, we fine-tune it on paired images from the PartDrag-4D and Objaverse-Animation-HQ datasets using its official checkpoint. For \textcolor{orange}{Puppet-Master}, we use its official checkpoint and test it only on the PartDrag-4D dataset.
In our setup, we set the input drag to the one with the maximum strength and sample frames at the 2\textit{nd}, 5\textit{th}, 8\textit{th}, 11\textit{th}, and 13\textit{th} positions to represent different strengths of drags.

% \noindent\textbf{Settings.} For the purpose of 3D reconstruction, we require multi-view images subjected to drag-deformation, which can be achieved through two distinct approaches. The first approach, referred to as \textit{NVS-First}, involves initially generating multi-view images using Zero123++, followed by applying drag-deformation (using ground truth multi-view 2D drags) to each individual view. The second approach, known as \textit{Drag-First}, applies drag-deformation directly to the input image before performing multi-view image synthesis via Zero123++. For 3D Gaussian Splattings reconstruction, we also leverage LGM to regress the gaussian representation for saving inference time, reported in Table~\ref{tab:sota}. 
% However, we also report the optimization-based inference time in Figure~\ref{fig:quati}.
\noindent\textbf{Settings.} 
We present a \textbf{novel evaluation approach} for part-level motion learning in 3D through the synthesis of novel views. While current methods typically generate single view's results, obtaining accurate 3D Gaussian outputs requires multi-view inputs. To address this gap, we propose two distinct approaches: NVS-First and Drag-First. The NVS-First approach involves first generating multi-view images using Zero123++, followed by the application of drag-deformation to each individual view. In contrast, the Drag-First approach applies drag-deformation directly to the input image before synthesizing multi-view images through Zero123++. 
%zero123++即使做了nvs，我们最终需要的是一个能在3d任意视角渲染的repr，之前的方法为了拿到这个repr，用optimzation-based training，我们把时间汇报到了表里和表里。为了对比reconstruction model本身的速度，
Although we have performed NVS via Zero123++, what we ultimately need is a 3D representation that can render from arbitrary viewpoints. Previous methods used optimization-based method to obtain this representation, where the time was reported in Figure~\ref{fig:quati} and ~\ref{tab:sota}. To align the reconstruction process with PartRM and reduce inference time, we leverage LGM to regress the Gaussian representation. The quality of the generated 3D model is evaluated using metrics, including PSNR, LPIPS, and SSIM, calculated based on rendered images of the Gaussians produced by this method.
% For 3D Gaussian Splatting reconstruction, we utilize LGM to regress the Gaussian representation, significantly reducing inference time, as demonstrated in Table~\ref{tab:sota}. Additionally, we report optimization-based reconstruction times for 3D Gaussian Splatting, showing that our method reduces inference time compared to both 3D reconstruction approaches.
% Additionally, we record the \textbf{optimization-based} 3D Gaussian Splatting results based on generated multi-view deformed images in our qualitative analysis.

\noindent\textbf{Quantitative Comparisons.} We conduct a quantitative evaluation to assess the performance of all compared methods. As shown in Table \ref{tab:sota}, our method outperforms state-of-the-art approaches across all metrics. This improvement is attributed to our pipeline's ability to capture both motion and geometry simultaneously, providing a synergistic advantage while also reducing computation time. In contrast, methods such as DragAPart, DiffEditor, and Puppet-Master rely solely on 2D image information, often leading to deformations that are inconsistent with the 3D structure of the objects.

\noindent\textbf{Qualitative Comparisons.} We provide a comparison in Figure \ref{fig:teaser} and \ref{fig:quati}. For DragAPart and DiffEditor, the multi-view images are rendered from Gaussian Splattings obtained through the optimization-based method. As illustrated in Figure~\ref{fig:quati}, DiffEditor fails to perform correct part-level deformations. Although DragApart can handle part-level deformations for articulated objects, it often produces unnatural artifacts (e.g., the microwave's door in the top-right subplot). Additionally, for objects with more complex or diverse deformation patterns, DragAPart struggles to capture the motions effectively. For Puppet-Master, we present both the originally generated images and those produced by Zero123++ in Figure \ref{fig:teaser}. While Puppet-Master maintains temporal consistency as drag strengths increase, it struggles to ensure precise motion accuracy (evident in the top left of Figure \ref{fig:teaser}) and multi-view consistency (as shown in the bottom left of Figure \ref{fig:teaser}). These limitations hinder its overall performance.

\begin{figure}
\centering
\includegraphics[width=0.35\textwidth]{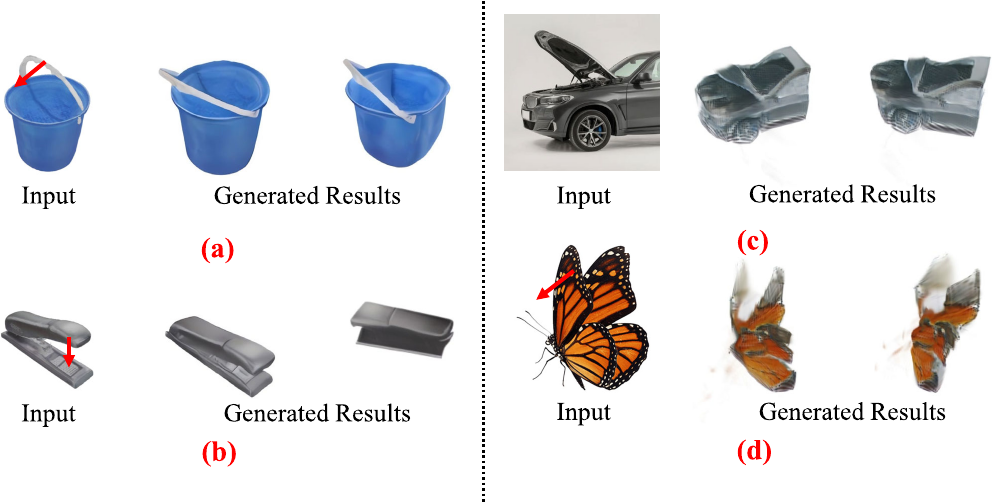}
\caption{\textbf{Generalization to the data in the wild}. We collect data from the internet for evaluation.}

% ~\eric{The motion in the dexterous hand example is not very obvious}
% }
\label{fig:real}
\vspace{-0.4cm}
\end{figure}

\noindent\textbf{Qualitative Results on In-the-Wild Data.} To evaluate the generalization capability of our method, we collected various articulated objects from the Internet and manually defined the drag configurations. The results are shown in Figure \ref{fig:real}, which indicate that PartRM generalizes well to motions close to the training distribution ((a) and (b)) but struggles with inputs data that deviate significantly ((c) and (d)).

\subsection{Ablation Studies}

\noindent\textbf{Ablation on Number of Drags.} To illustrate the effectiveness of our drag-propagation module, we performed ablation studies on the number of drags using the PartDrag-4D dataset. As shown in Table \ref{tab:ablation_drags}, the performance improves with an increasing number of drags. This improvement can be attributed to the fact that more input drags provide clearer guidance on moving parts, thereby enhancing the modeling of part motion dynamics.

\begin{table}[tbp]
    \centering
    \resizebox{0.8\columnwidth}{!}{%
        \begin{tabular}{cccc}
            \toprule
            \textbf{Num Drags} & \textbf{PSNR (↑)} & \textbf{SSIM (↑)} & \textbf{LPIPS (↓)} \\ 
            \midrule
            1 & 27.06 & 0.9466 & 0.0452 \\ 
            5 & \underline{27.56} & \underline{0.9483} & \underline{0.0448} \\ 
            10 & \textbf{28.15} & \textbf{0.9531} & \textbf{0.0356}  \\ 
            % 30 & \textbf{28.15} & \textbf{0.9531} & \textbf{0.0356} \\
            \bottomrule
        \end{tabular}
    }
    \caption{Ablation on number of drags.}
\vspace{-0.5cm}
    \label{tab:ablation_drags}
\end{table}

\begin{table}[tbp]
    \centering
    \resizebox{\columnwidth}{!}{%
        \begin{tabular}{cccc}
            \toprule
            \textbf{Training Stage Setting} & \textbf{PSNR (↑)} & \textbf{SSIM (↑)} & \textbf{LPIPS (↓)} \\ 
            \midrule
            Only w/ Stage 1 & 22.05 & 0.8624 & 0.1274 \\ 
            Only w/ Stage 2 & \underline{25.87} & \underline{0.9387} & \underline{0.0537} \\ 
            Stage 1+2 & \textbf{28.15} & \textbf{0.9531} & \textbf{0.0356}  \\ 
            \bottomrule
        \end{tabular}
    }
    \caption{Ablation on training stages.}
\vspace{-0.25cm}
    \label{tab:ablation_stage}
\end{table}

\begin{table}[tbp]
    \centering
    \resizebox{\columnwidth}{!}{%
        \begin{tabular}{cccc}
            \toprule
            \textbf{Drag Embeddings Setting} & \textbf{PSNR (↑)} & \textbf{SSIM (↑)} & \textbf{LPIPS (↓)} \\ 
            \midrule
            Only w/ 128*128 scale & 25.48  & \textbf{0.957}  & 0.048  \\ 
            Only w/ 32*32 scale & \underline{27.99}  &  0.952 & \underline{0.039}  \\ 
            Only w/ 8*8 scale & 26.87 & \underline{0.953} & 0.045 \\ 
            Multi-scale & \textbf{28.15} & \underline{0.953} & \textbf{0.036}  \\ 
            \bottomrule
        \end{tabular}
    }
    \caption{Ablation on multi-scale drag embeddings.}
\vspace{-0.4cm}
    \label{tab:ablation_multiscale}
\end{table}

\begin{figure}[t]
\centering
\includegraphics[width=0.85\linewidth]{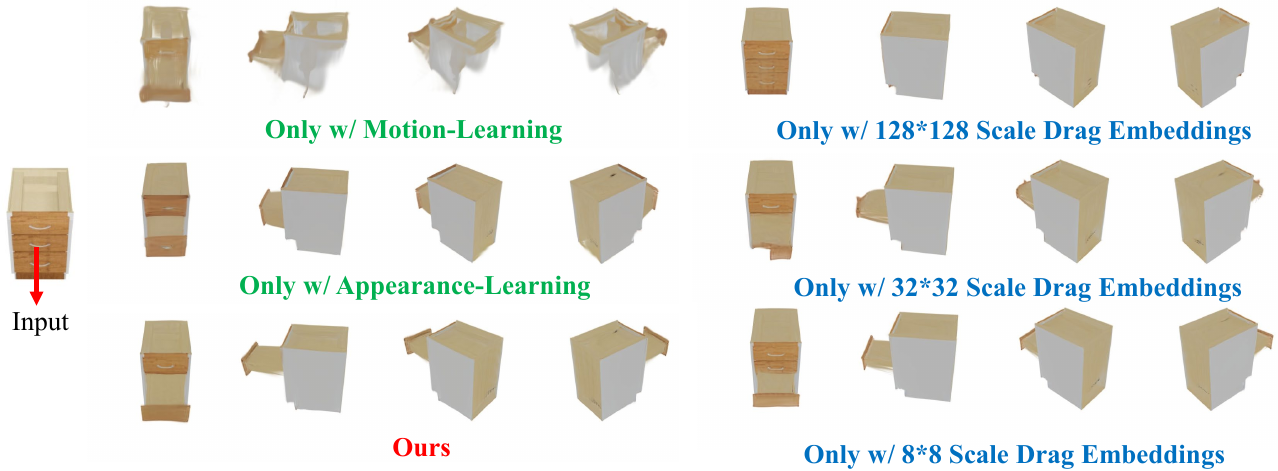}
\caption{\textbf{Ablation on the training stage and multi-scale drag embeddings.} 
}
\vspace{-0.6cm}
\label{fig:ablation}
\end{figure}

\noindent\textbf{Ablation on Training Stages.} To demonstrate the effectiveness of our two-stage training process, we conducted ablation studies on the PartDrag-4D dataset using 10 propagated drags. As shown in Table \ref{tab:ablation_stage}, employing only one training stage leads to a significant performance drop. The visualization results in Figure \ref{fig:ablation} further illustrate this point. When only the motion learning stage is used, the appearance quality is poor. Conversely, if only the appearance learning stage is employed, the drawer in the image fails to move to the correct position. This limitation arises from the weak supervision provided by multi-view images, which hinders the effective continual learning of motion dynamics over the learned appearance and geometry modeling.

\noindent\textbf{Ablation on Muti-scale Drag Embeddings.} To validate the effectiveness of our multi-scale drag embeddings, we conduct ablation studies on embedding scales using the PartDrag-4D dataset with 10 propagated drags. In this experiment, we concatenate drag embeddings to the UNet downsample blocks with spatial dimensions of 8×8, 32×32, and 128×128. As shown in Table \ref{tab:ablation_multiscale}, our multi-scale embedding design achieves higher PSNR and lower LPIPS. As illustrated in the right panel of Figure \ref{fig:ablation}, concatenating only large spatial dimension (128×128) drag embeddings to the UNet impairs motion learning, as this scale lacks the capacity to capture the global features of moving parts. On the other hand, concatenating only small spatial dimension (8×8) drag embeddings will mislead the model about which part to move. Specifically, as noted in Sec. \ref{sec:drag_embedding}, the drag intended to open a drawer in the second layer may overlap spatially on the drag map with the drag intended to open the last layer’s drawer, leading to ambiguity. Although using an intermediate spatial scale partially mitigates these issues, it does not achieve optimal performance.

% \subsection{Real World Data Results}

\subsection{Sim2Real Applications in Manipulation}
\label{sec:embodied}

\begin{figure}
\centering
\includegraphics[width=0.8\linewidth]{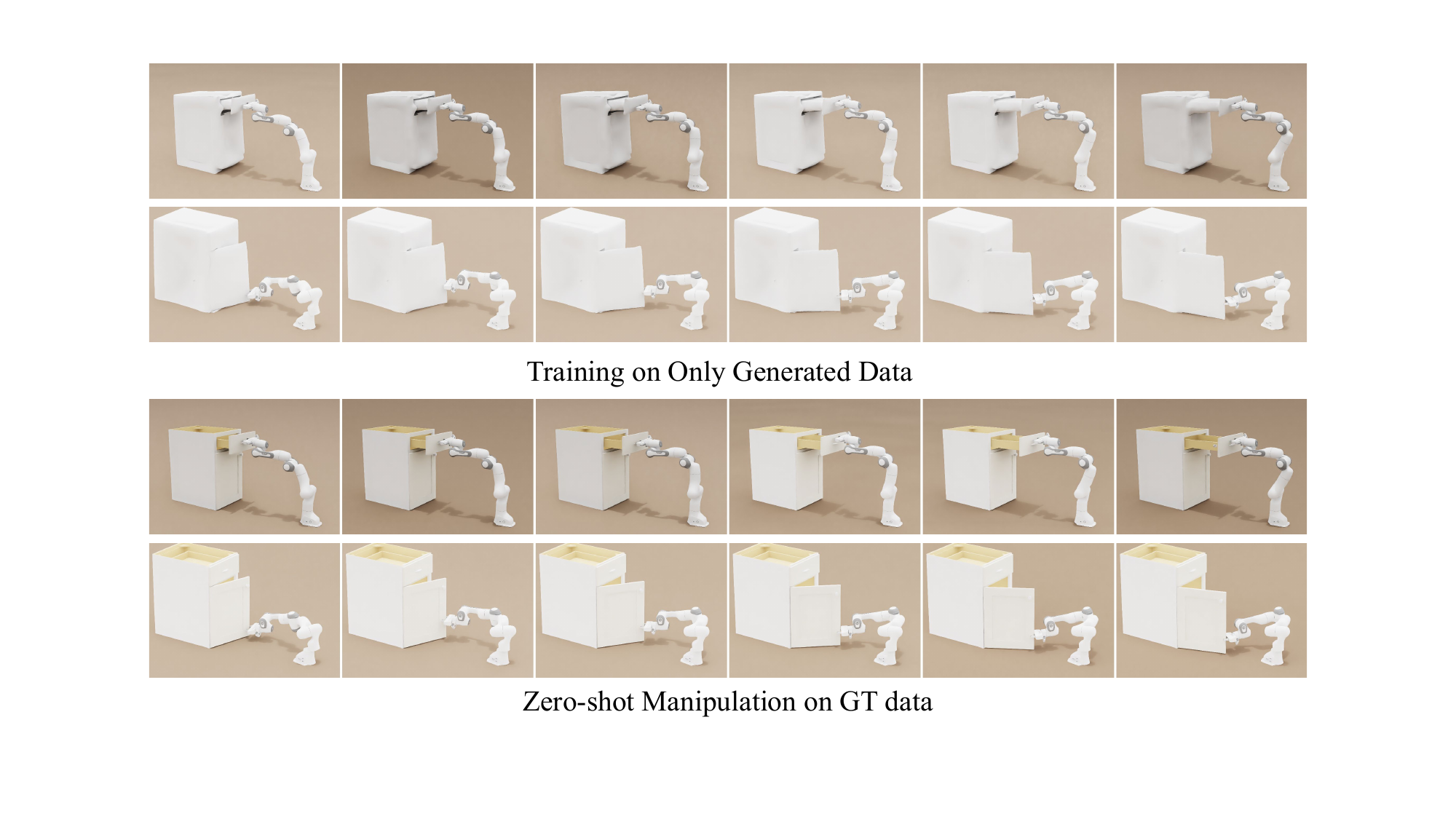}
\caption{\textbf{Applications in robot manipulation.} We get the articulated object manipulation policy on our generated data and generalize it to the ground-truth data.
}
\vspace{-0.6cm}
\label{fig:embodied}
\end{figure} 
To assess the applicability of our method in robotic manipulation, we conduct a sim-to-real experiment using the Isaac Gym~\cite{makoviychuk2isaac} simulator. In this experiment, a manipulation policy is trained using data generated by our approach, and its ability to generalize to real-world ground-truth data is subsequently evaluated. We are curating that successful generalization to real-world data demonstrates that our generated data can be effectively utilized to train a manipulation policy with only a single-view image of an object instead of predicting affordance~\cite{chen2022cerberus, li2022toist}. Specifically, given a single-view input and corresponding 2D drags on the selected part by the users, we employ PartRM to generate both the initial and deformed states. These states are then used to extract the part meshes of the moving objects and their corresponding axes of motion ~\cite{zhong20233d}, facilitating the training of the policy.

We design an experimental task involving the manipulation of two distinct objects: a prismatic drawer and a rotational door. The objective is to move each object from a partially opened state to a fully opened state. Using the part meshes and axes extracted from our generated data, we can get the manipulation policy in Issac Gym \cite{makoviychuk2isaac} to accomplish this task. The policy is then evaluated to assess its ability to successfully operate in a \textit{ground-truth URDF} environment. As shown in Figure \ref{fig:embodied}, the Franka robot arm can conduct zero-shot manipulation on the ground-truth data, effectively completing the specified task. More details of this experiment are in the \href{#sec:Appendix}{Appendix}.

% Given an input image and corresponding 2D drags, we can use our model to generate the initial and deformed states. We can then extract the part meshes of the moving parts and the corresponding axis.
% % , which can serve as training data for robotic manipulation tasks on articulated objects. 
% Given the current scarcity of human-annotated articulated data, the generated data will be highly beneficial for advancing the field of embodied AI.

% To evaluate the efficacy of the generated data, we design an experimental task involving the manipulation of two distinct objects: a prismatic drawer and a rotational door. The objective is to move each object from a partially opened state to a fully opened state. Using the part meshes and axes extracted from our generated data, we can get the manipulation policy in Issac Gym \cite{makoviychuk2isaac} to accomplish this task. The policy is then evaluated to assess its ability to successfully operate in a \textit{ground-truth URDF} environment. As shown in Figure \ref{fig:embodied}, the Franka robot arm can generalize the policy to the ground-truth data, effectively completing the specified task. This demonstrates that our generated data aligns well with the ground truth and is of high quality. The details of this experiment are in the \href{#sec:Appendix}{Appendix}.
\vspace{-0.2cm}
\section{Conclusions}
This paper introduces PartRM, a novel approach that simultaneously models appearance, geometry, and part-level motion. To address data scarcity in 4D part-level motion learning, we present the PartDrag-4D dataset, which provides multi-view observations of part-level dynamics. Experimental results show that our method outperforms previous approaches in part motion learning and is applicable to embodied AI tasks. However, it may struggle with articulated data deviating significantly from the training distribution. We believe our insights and the high-quality images generated by our model will inspire future research.

\section*{Acknowledgments} This work is supported by the National Natural Science Foundation of China (NSFC) under Grants No. 623B1010.

{
    \small
    \bibliographystyle{ieeenat_fullname}
    \bibliography{main}
}
\clearpage
\newpage

\vspace{-5.0cm}
\section{Additional Details}
\label{sec:Appendix}
% \subsection{Toolkit}
% For access to our anonymous code toolkit, please visit: \href{https://anonymous.4open.science/r/PartRM-7C81/}{https://anonymous.4open.science/r/PartRM-7C81/}
\subsection{Details of PartDrag-4D Dataset}

\textit{Category Details}: We extract eight object categories from PartNet-Mobility \cite{xiang2020sapien}: dishwasher, laptop, microwave, oven, refrigerator, storage-furniture, washing-machine, and trashcan. Notably, the trashcan category is excluded from the training phase, serving as an unseen category. The extraction of specific categories is guided by the PartNet-Mobility dataset's detailed part-level annotations.

\textit{Rendering Details}:
We utilize Blender 3.5.0 for rendering multi-view images. The camera is fixed at a distance of 2.4 meters and a height of 1.5 meters, with 12 views rendered uniformly across the scene. The rendered images are produced at a resolution of 512 × 512 resolutions and stored in RGBA format. Both the images and their corresponding camera parameters are saved for subsequent processing.

\subsection{Details of Training and Evaluation}
\textit{Training Details}: Our model was trained on a setup consisting of 4 × NVIDIA A800 GPUs, with a batch size of $4 \times 8$ and a learning rate of $5  \times 10^{-4}$. The training process included 100,000 iterations for the motion-learning phase and 50,000 iterations for the appearance-learning phase, utilizing the AdamW optimizer.

\textit{Evaluation Details}: For evaluation, we randomly sample an initial state (e.g. a partially opened drawer) and apply drag-based deformation to transition to other states. This enables us to assess whether our model, as well as the baselines, can accurately perform the deformations. 
% For DiffEditor \cite{mou2024diffeditor}, we specifically utilize its \textit{drag} mode during the evaluation process.

\subsection{Details of Applications in Manipulation Task}

\textit{Details of Mesh and Axis Extraction}: To facilitate robotic manipulation in Isaac Gym \cite{makoviychuk2021isaac}, the object's mesh is first derived by extracting it from generated Gaussian Splattings \cite{kerbl3Dgaussians} as detailed in LGM \cite{tang2025lgm}. Leveraging the meshes of these two states, the moving part and its axis can be extracted following \cite{zhong20233d} by uniformly sampling 10,000 points from the meshes as input.

% After obtaining meshes for two states, we extract the axis of motion following the method in \cite{todo 3d implicit transporter}, with input point clouds sampled from the generated meshes. 
% After getting the manipulation Finally, we test whether the policy can generalize to real-world settings represented by the URDF, thereby evaluating if our generated data aligns well with the ground truth.

\section{Additional Experiment Results}
\subsection{Ablation Study on Drag Propagation Method}
\begin{figure}[h]
\centering
\includegraphics[width=\linewidth]{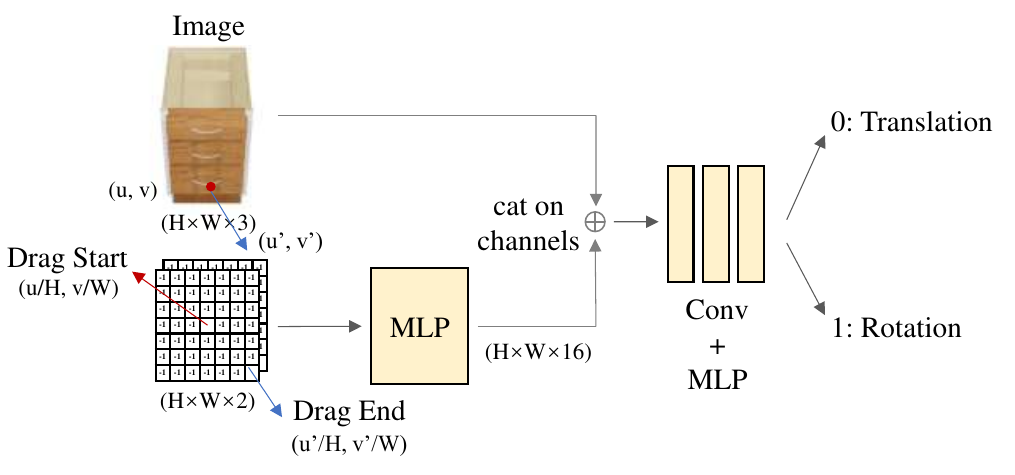}
% \vspace{-0.5cm}
\caption{Drag classification network structure.}
\vspace{-0.3cm}
\label{fig:rebuttal_drag}
\end{figure}

The drag propagation module is designed to enable PartRM to better understand the regions of moving parts. PartRM learns the translation and rotation of these parts utilizing a large amount of data, which can be considered as a data-driven approach. To inspect whether the rule-based propagation module can have a better performance, we have developed a pipeline that incorporates a drag classification model which takes in the input images and drags
(Fig~\ref{fig:rebuttal_drag}) to categorize the type of drag (i.e., translation or rotation), coupled with a rule-based method for drag propagation. Given input drags parameterized by its start and end points projection, i.e., $a_t = (a_{t,src}(x, y), a_{t, dst}(x,y))$. We denote $\Delta a_{t} = a_{t,dst} - a_{t,src}$ and the i-th propagated drag as $a_{t, i}$. For translation, our propagated rules can be formulated as:
\begin{equation}
    a_{t,i} = (a_{t,i,src}, a_{t,i,src} + \Delta a_{t})
\end{equation}
where $a_{t,i,src}$ is the i-th point sampled from segmentation mask generated by Segment Anything~\cite{kirillov2023segment}. For rotation, we define the propagated rules:
\begin{equation}
    \begin{aligned}
        a_{t,i} &= \left( a_{t,i,\text{src}}, \right. \\
        &\left. a_{t,i,\text{src}} + \Delta a_t \left( 1 - \frac{\Delta a_t * \left( a_{t,i,\text{src}} - a_{t,\text{src}} \right)}{\underset{a_{t,j,\text{src}}}{\max} \Delta a_t * \left( a_{t,j,\text{src}} - a_{t,\text{src}} \right)} \right) \right)
    \end{aligned} 
\end{equation}
where $a_{t,j,\text{src}}$ is the point sampled from the part segmentation mask, and $*$ represents the inner product of two vectors. 

From our experiment, the drag classification model achieves \textbf{69.1\%} accuracy on the PartDrag-4D test set.
We also conduct an ablation study on the propagation methods, as shown in Table~\ref{tab:ablation_prop}. The results demonstrate that PartRM outperforms this new method by effectively capturing both translational and rotational deformations through the synergistic learning of geometry and drag deformation.

\begin{table}[tbp]
    \centering
    \resizebox{\columnwidth}{!}{%
        \begin{tabular}{cccc}
            \toprule
            \textbf{Prop. Method} & \textbf{PSNR (↑)} & \textbf{SSIM (↑)} & \textbf{LPIPS (↓)} \\ 
            \midrule
            Rule-Based & 27.04 & 0.9368 & 0.0402\\ 
            Data-Driven (Ours) & \textbf{28.15} & \textbf{0.9531} & \textbf{0.0356}  \\ 
            \bottomrule
        \end{tabular}
    }
    \caption{Ablation on propagation methods.}
    \label{tab:ablation_prop}
\end{table}

\subsection{Ablation Study on NVS Method}
To assess the impact of the novel view synthesis (NVS) methods, we utilize an alternative NVS technique \cite{kong2024eschernet}, while maintaining all other experimental conditions constant. As demonstrated in Table~\ref{tab:ablation_nvs}, the application of this more advanced NVS method results in a modest improvement in performance.

\begin{table}[tbp]
    \centering
    \resizebox{\columnwidth}{!}{%
        \begin{tabular}{cccc}
            \toprule
            \textbf{NVS Method} & \textbf{PSNR (↑)} & \textbf{SSIM (↑)} & \textbf{LPIPS (↓)} \\ 
            \midrule
            EscherNet & \textbf{28.22} & \textbf{0.9574} & \textbf{0.0352} \\ 
            Zero123++ (Ours) & 28.15 & 0.9531 & 0.0356  \\ 
            \bottomrule
        \end{tabular}
    }
    \caption{Ablation on NVS methods.}
\vspace{-0.3cm}
    \label{tab:ablation_nvs}
\end{table}

\subsection{Ablation Study on Drag Embedding Method}

To assess the impact of various drag embedding methods, we conducted an ablation study focusing on the drag embedding techniques and injection approaches described in DragAPart~\cite{li2025dragapart} and Puppet-Master~\cite{li2024puppet}. All other experimental conditions were kept consistent with those in PartRM. Notably, neither DragAPart nor PartRM utilizes Fourier Embedding for encoding input drags. In DragAPart, the encoded drags are directly concatenated along the channels of the UNet features, whereas Puppet-Master divides the encoded drags into two blocks, treating them as the scale and shift parameters to apply on the UNet feature.
As demonstrated in Table~\ref{tab:ablation_drag_embedding}, PartRM outperforms other methods, with the incorporation of Fourier embeddings further enhancing its performance.

\begin{table}[tbp]
    \centering
    \resizebox{\columnwidth}{!}{%
        \begin{tabular}{cccc}
            \toprule
            \textbf{Embedding Method} & \textbf{PSNR (↑)} & \textbf{SSIM (↑)} & \textbf{LPIPS (↓)} \\ 
            \midrule
            DragAPart & 27.51 & 0.9507 & 0.0403 \\ 
            Puppet-Master & 27.74 & \textbf{0.9545} & 0.0411  \\ 
            PartRM (Ours) & \textbf{28.15} & 0.9531 & \textbf{0.0356} \\ 
            \bottomrule
        \end{tabular}
    }
    \caption{Ablation on Drag Embedding methods.}
\vspace{-0.3cm}
    \label{tab:ablation_drag_embedding}
\end{table}

\subsection{Qualitative Study on the Ambiguity of Drags}

\begin{figure}[h]
\centering
\includegraphics[width=0.7\linewidth]{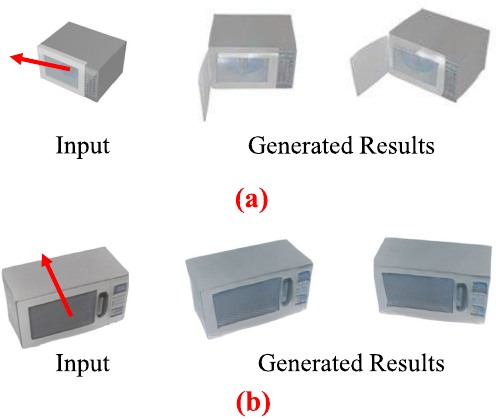}
% \vspace{-0.5cm}
\caption{Qualitative Study on the Ambiguity of Drags.}
\vspace{-0.3cm}
\label{fig:final_drag}
\end{figure}

To inspect whether PartRM can handle the drags ambiguity well, we conduct a qualitative study as shown in Figure~\ref{fig:final_drag}. PartRM can handle minor perturbations ((a)) due to noise introduced during training. 
When drags significantly differ from the training data, the model fails to produce the expected outcome (b), where the drags attempt to push the entire microwave inside (the microwave's fully closed door cannot be articulated). 
This is because we don't model the motion of the whole object (only part motion) and there is lack of related training data. 

\section{Additional Visualization Results}
\begin{figure*}
\centering
\includegraphics[width=\textwidth]{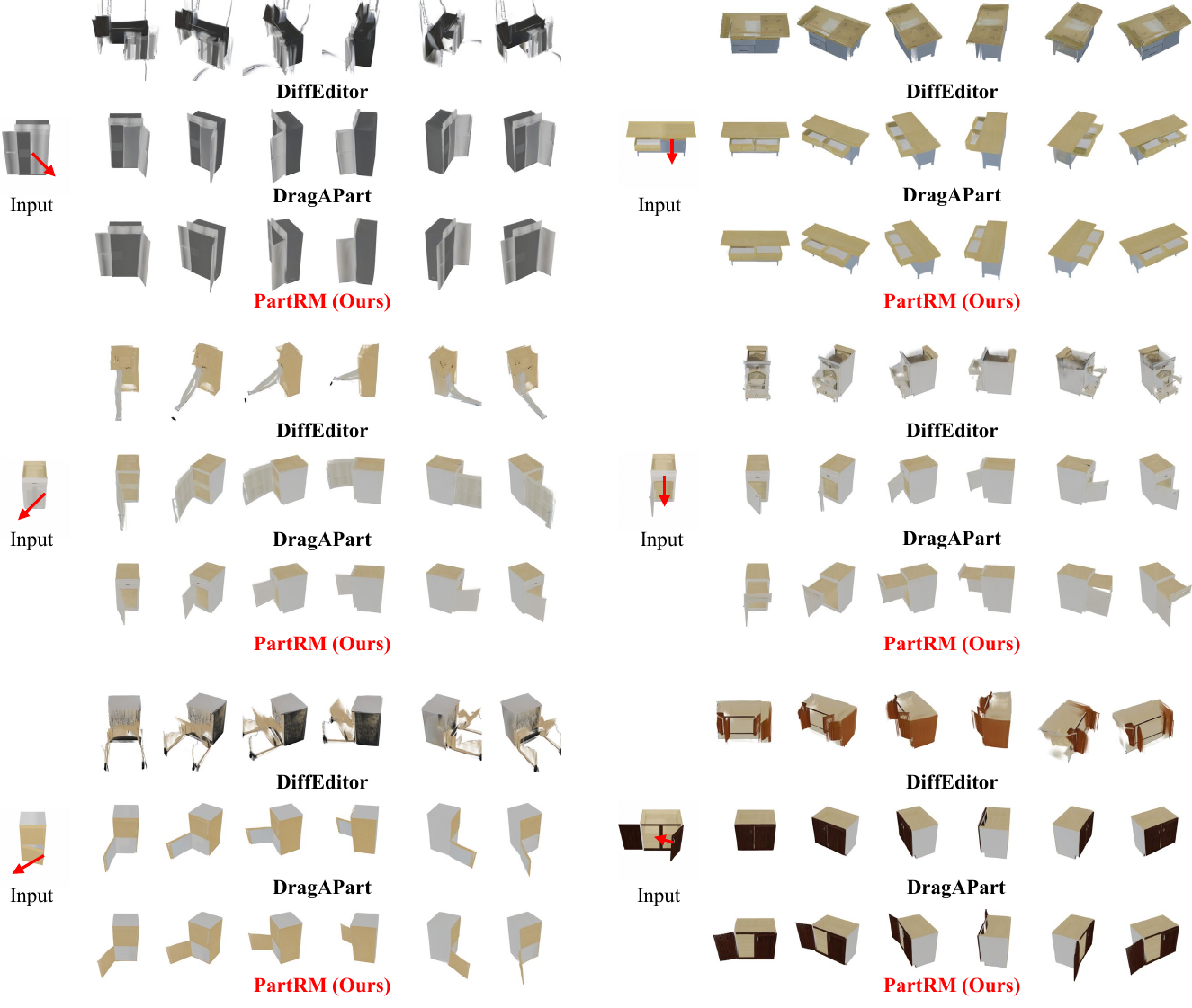}
\vspace{0.3cm}
\caption{Qualitative comparisons between PartRM and baselines on PartDrag-4D dataset.
}
\label{fig:supp_part}
\end{figure*}

\begin{figure*}
\centering
\includegraphics[width=\textwidth]{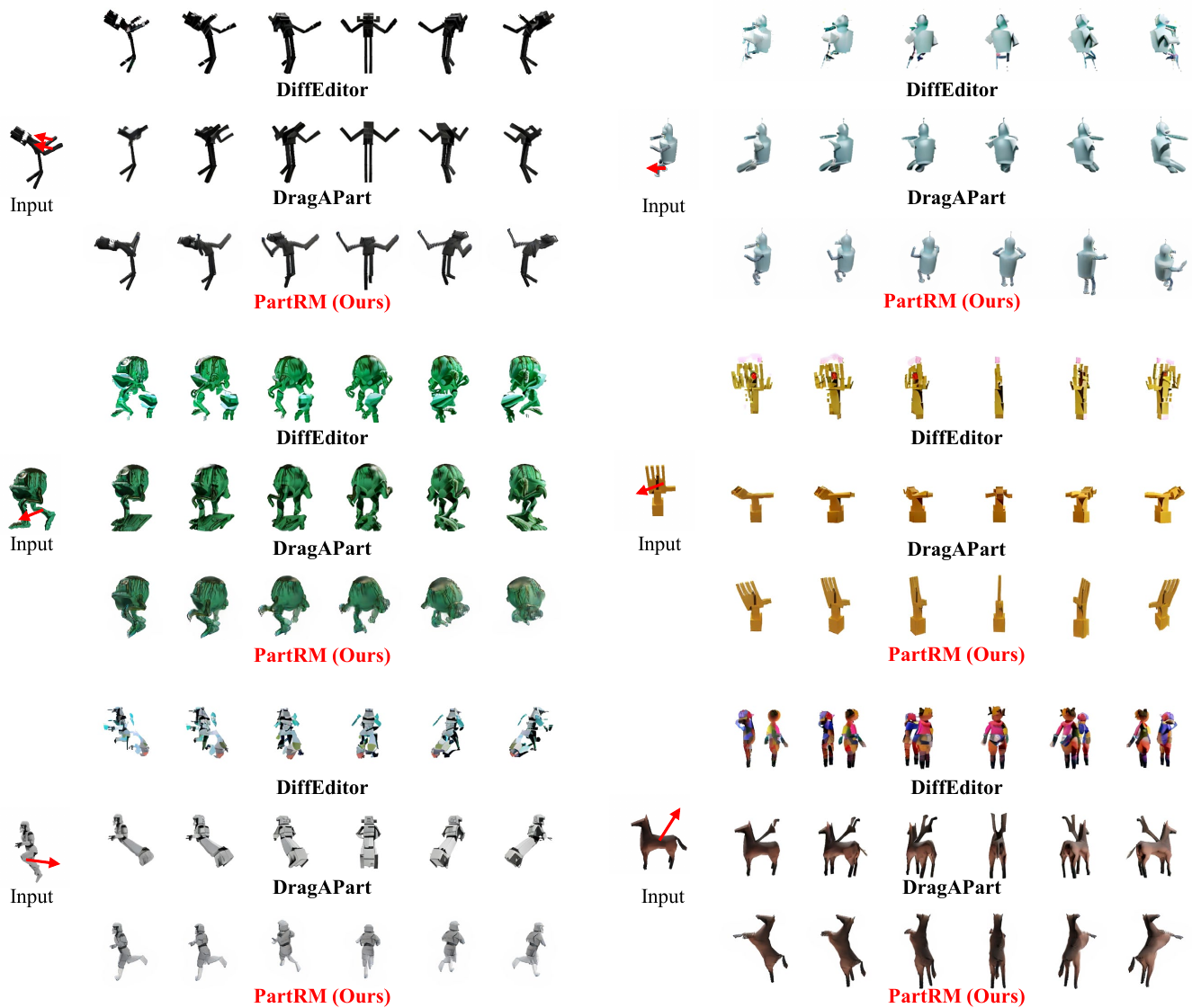}
\vspace{0.3cm}
\caption{Qualitative comparisons between PartRM and baselines on Objaverse-Animation-HQ dataset.
}
\label{fig:supp_obj}
\end{figure*}

\begin{figure*}
\centering
\includegraphics[width=\textwidth]{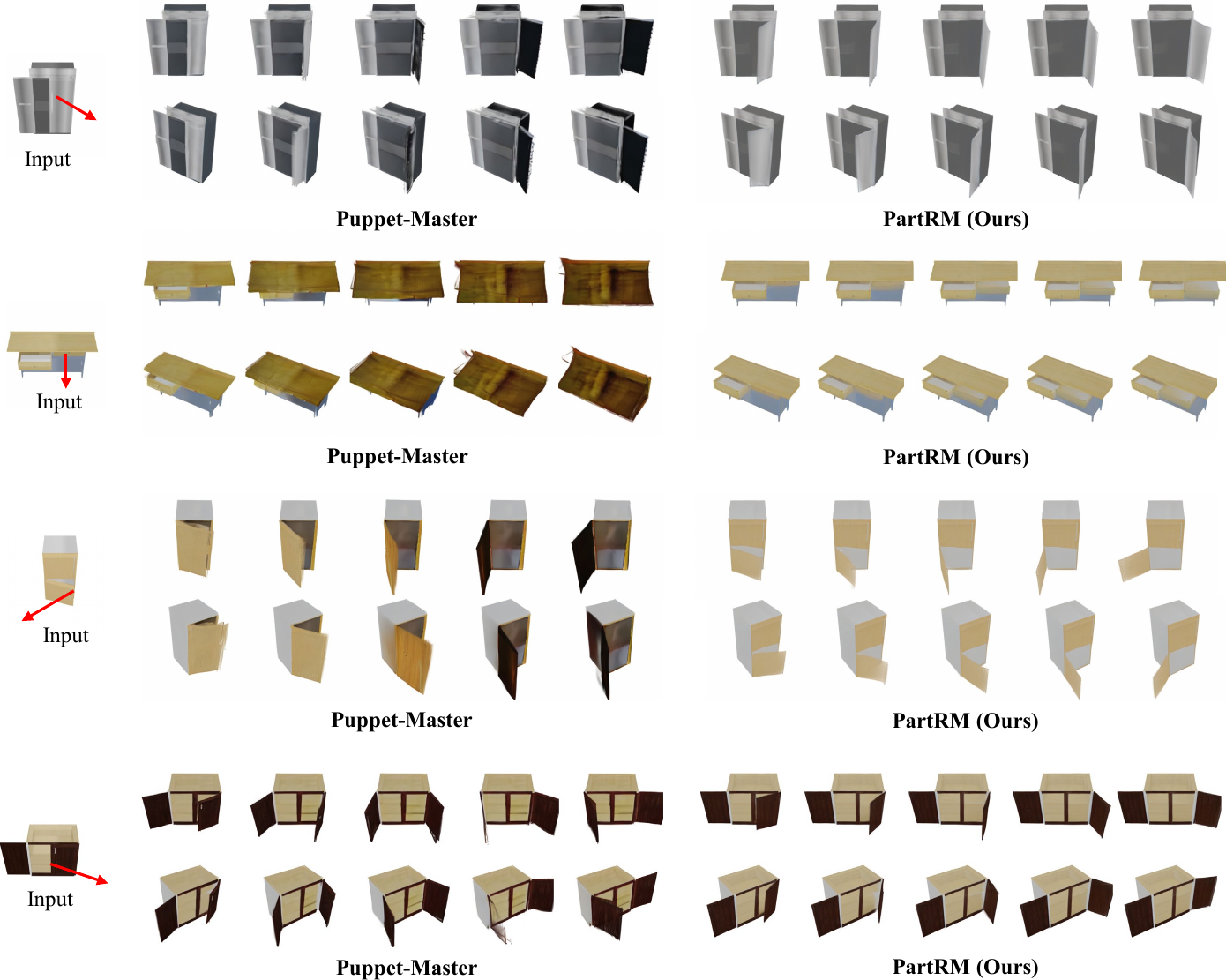}
\vspace{0.3cm}
\caption{Qualitative comparisons between PartRM and Puppet-Master on PartDrag-4D dataset.
}
\label{fig:supp_pm}
\end{figure*}

We provide more qualitative results. Please refer to Figure \ref{fig:supp_part}, Figure \ref{fig:supp_obj} and Figure \ref{fig:supp_pm}
\label{sec:more_results} for details.

% \input{sec/X_supple}

% {
%     \small
%     \bibliographystyle{ieeenat_fullname}
%     \bibliography{main}
% }

% \newpage

% {
%     \small
%     \bibliographystyle{ieeenat_fullname}
%     \bibliography{main}
% }

% \input{sec/X_supple}

\end{document}